# Where Classification Fails, Interpretation Rises

**Chanh Nguyen, Georgi Georgiev, Yujie Ji, Ting Wang**
Department of Computer Science and Engineering
Lehigh University
Bethlehem, PA 18015
{cpn217, gdg217, yuj216, ting}@cse.lehigh.edu

## Abstract

An intriguing property of deep neural networks is their inherent vulnerability to adversarial inputs, which significantly hinders their application in security-critical domains. Most existing detection methods attempt to use carefully engineered patterns to distinguish adversarial inputs from their genuine counterparts, which however can often be circumvented by adaptive adversaries. In this work, we take a completely different route by leveraging the definition of adversarial inputs: while deceiving for deep neural networks, they are barely discernible for human visions. Building upon recent advances in interpretable models, we construct a new detection framework that contrasts an input's interpretation against its classification. We validate the efficacy of this framework through extensive experiments using benchmark datasets and attacks. We believe that this work opens a new direction for designing adversarial input detection methods.

## 1 Introduction

Recent advances in deep learning have led to breakthroughs in long-standing artificial intelligence tasks, e.g., image classification, speech recognition, and game playing, and enabled use cases previously considered strictly experimental. Yet, deep neural networks (DNNs) are inherently vulnerable to adversarial inputs [1], those maliciously crafted samples to trigger DNNs to misbehave, which significantly hinders DNN's application in security-critical domains.

Since the discovery of such vulnerabilities [1], a variety of attack models have been proposed. For example, Jacobian Saliency Map Attack [2] iteratively picks pixels ($l_0$) and perturbs them according to their effect on achieving misclassification; L-BFGS [1], DeepFool [3], Universal [4], C&W [5] attacks are all using Euclidean (root-mean-square) method ($l_2$) to measure the influence of perturbations; while Fast Gradient Sign Method [1] and Basic Iterative Method [6] change every pixel of the original image simultaneously ($l_\infty$). See the Appendix for adversarial samples generated by each type of attacks.

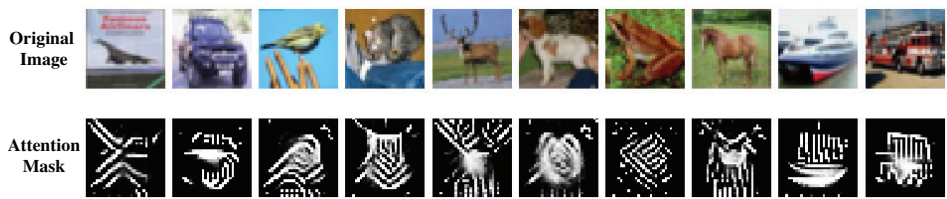

Figure 1: Original images from each class and their masks generated by LAN.

On the other hand, a plethora of defense mechanisms has been proposed. The existing methods can be roughly categorized into two classes: one that reduces the influence of distortion on the model's inputs [7], [8], [9], [10] and the other utilizes the model's outputs to help make more robust classification [7], [11], [12], [9], [10]. Yet, relying on carefully engineered patterns to distinguish genuine and adversarial inputs, most of the defenses can often be circumvented by adaptive adversaries or new attack variants [13].

In this paper, we propose a new detection framework that completely departs from existing efforts. Intuitively, we revisit the fundamental definition of adversarial inputs, which are examples that can deceive neural networks but not humans, because humans have the ability to extract the main information from an image and ignore adversarial perturbations. We try to mimic this ability by leveraging attention mechanisms to generate representative patterns for each class of the input images. Specifically, our work is inspired by Latent Attention Network (LAN) [14] which, for each input example, generates a mask, called "attention mask", that represents the most important pixels of that image when a network tries to classify it. Figure 1 shows images and their corresponding attention masks. The idea of our detection method is as follows: the attention mask of an adversarial image remains similar to that of its corresponding benign even when the image can fool the target classifier.

As far as we know, we're the first to apply an attention mechanism to the adversarial examples detection. We test our method against state-of-the-art attacks and we show promising initial results following this idea. We also look into cases where our approach fails and then point out potential directions for future research.

## 2 Method

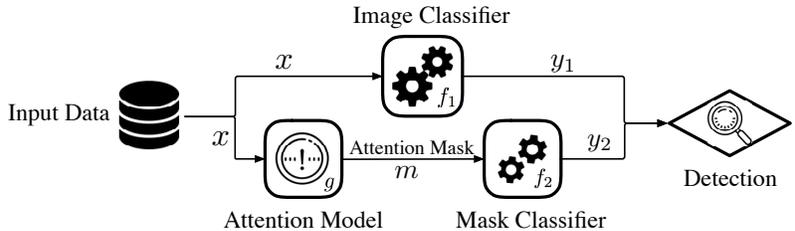

Figure 2: Framework of classification-interpretation contrastive detection.

As shown in Figure 2, our detection method is characterized by three modules: an image classifier $f_1$, an attention model $g$, and a mask classifier $f_2$. The image classifier $f_1$ is a function $F_1 : \mathbb{R}^d \to \mathbb{R}^l$ which, given input $x$, classifies it into decision $y_1$. Attention model $g$ is a LAN, which is a function $G : \mathbb{R}^d \to [0, 1]^d$ that, given an input $x$, produces an attention mask $m = G(x)$. Attention mask $m$ determines the important components of $x$ that influence the classification output of a classifier $f$ by corrupting pixels of $x$ with noise drawn from a predefined distribution and measures the change in $f$'s loss. The larger the loss is, the more important the pixels are. The resulting masks can capture the common features in images of the same class. $f$ can be any common classifier for $x$. In our experiments, we directly use $f_1$. The mask classifier $f_2$ is a function $[0, 1]^d \to \mathbb{R}^l$ which, given a mask $m$, classifies it into decision $y_2$. If $y_2$ agrees with $y_1$, we decide the image is benign and adversarial otherwise.

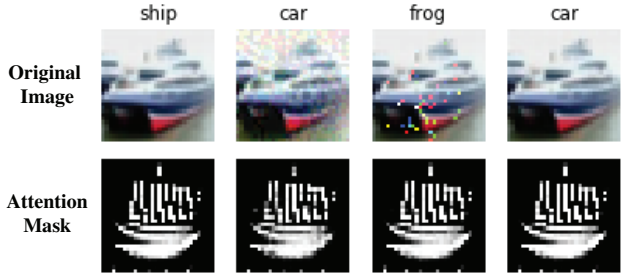

Figure 3: First row shows one benign image and its three adversarial examples by FGSM, JSMA, and C&W together with their classification results. The second row presents their corresponding attention masks. Visually, the masks look really similar to one another despite adversarial pertubations.

## 3 Evaluation

We experiment on CIFAR10 [15], with 50,000/10,000 train/test split. We use the same architecture in [14] for $f_1$ and $g$, which are AlexNet [16] and a 3-layer Fully Connected Network. $f_2$ is based on LeNet [17] and trained on the masks produced by $g$ from the training set. We test our method against three attacks corresponding to different distance metrics: $l_\infty, l_0, l_2$, namely FGSM, JSMA and C&W. In the rest of the paper, adversarial examples are treated as the positive class and we use $x$, $m$, and $x^*$, $m^*$ to denote benign images, attention masks, and their adversarial counterparts, respectively.

### 3.1 Invariance of mask

Our first experiment evaluates the applicability of attention masks to detecting adversarial samples. We generate attention masks for benign and malicious images using a LAN. Our intuition that attention masks of both types of images are very similar is confirmed and the results are shown in Figure 3. We also give a statistics analysis on how the mask classification results will change after perturbations. As shown in Table 1, over 85% of the adversarial samples fail to change their mask classifications, no matter how they are generated. This discovery and the quality of LANs to generate similar masks for images from the same class motivate our detection framework.

Table 1: Percentage of adversarial samples whose attention masks retain their original classifications

| Attack | Percentage |
|---|---|
| FGSM ($l_\infty$) | 0.863 |
| JSMA ($l_0$) | 0.878 |
| C&W ($l_2$) | 0.997 |

### 3.2 Effectiveness of detection

In the second experiment, we evaluate the effectiveness of the detection framework by comparing the prediction for image $x$ from classifier $f_1$ against the classification of the mask $m$ of image $x$. If they differ, we predict adversarial, and benign otherwise. For each attack method, we take the adversarial examples $x^*$ that successfully fool $f_1$. We pair the same amount of benign images with adversarial examples to create a test set. The results of our solution against FGSM, JSMA and C&W are shown in Table 2.

Adversarial examples are known to be able to transfer across models, so we also test our detection framework in a transferred setting. Table 3 shows the results of our method against adversarial samples generated for VGG-like, a modified VGG network [18] that has better accuracy than our AlexNet model.

Table 2: Classification accuracy on adversarial samples generated using different attacks on AlexNet

| Attack | True positive | True negative |
|---|---|---|
| FGSM ($l_\infty$) | 0.878 | 0.614 |
| JSMA ($l_0$) | 0.960 | 0.614 |
| C&W ($l_2$) | 0.860 | 0.614 |

Table 3: Classification accuracy on adversarial samples generated with different attacks on VGG-like

| Attack | True positive | True negative |
|---|---|---|
| FGSM ($l_\infty$) | 0.843 | 0.665 |
| JSMA ($l_0$) | 0.853 | 0.647 |
| C&W ($l_2$) | 0.917 | 0.750 |

Overall, our method shows good true positive rates: being able to detect adversary when the input image is indeed adversarial, in both direct and transferred setting. However, the true negative rates across different attacks are low, mostly under 70%: the mask classifier $f_2$ is confused when presented with a benign example. To figure out why, we calculate the accuracy of $f_2$ on the produced masks of the test set and we get 60%, which is quite low, compared to the accuracy of 82% of $f_1$, which classifies raw input images instead of the derived attention masks. Looking into the 40% of the masks that got incorrectly classified by $f_2$, we find the problem: those masks are mostly not of the common form of the masks of their corresponding classes. Figure 4 shows 2 examples.

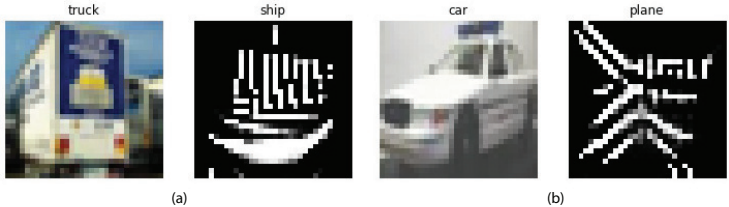

Figure 4: (a) and (b) show two examples where LAN produces attention masks with a totally different class from their corresponding original images.

### 3.3 Relability of detection

We then study the quality of the mask generator network and see how it affects our detection method's performance. We filter out the masks that are incorrectly produced by LAN and are left with the 60% of the test set, and generate adversarial samples for those cases. We repeat our experiments with this smaller data set and get the following results in Table 4. We achieve the exactly same results for AlexNet and VGG-like. Overall, when the masks for benign images are correctly produced by LAN, our detection accuracy is perfect across all the attack methods, in both direct and transferred settings. What's more, the recovery rates (retrieving the original classification of the network despite adversarial perturbations) are also 100%.

Table 4: Classification accuracy on adversarial samples generated using different attacks on AlexNet, after filtering out the incorrect masks.

| Attack | True positive | True negative | Recovery rate |
| --- | --- | --- | --- |
| FGSM ($l_\infty$) | 1.000 | 1.000 | 1.000 |
| JSMA ($l_0$) | 1.000 | 1.000 | 1.000 |
| C&W ($l_2$) | 1.000 | 1.000 | 1.000 |

## 4 Conclusion and Discussion

We propose an attention-based framework for defending against adversarial examples, which is an initial step to utilize a model's interpretability. Our method uses an attention mask generator, specifically a LAN, to find an input image representation, which is invariant regardless of adversarial modifications. We show that attention masks are resilient against adversarial perturbations and build our adversary detection based on that property. Our initial experiments provide promising results with a good detection performance. The framework's perfect detection accuracy and recovery rates, after filtering out benign images with incorrect masks, hint at a potential increase in detection accuracy if we can optimize the quality of the attention mask generator. The proposed method is also attack-agnostic in that it does not need to know the specifics in adversarial samples generation process. However, our detection method's performance is highly dependent on the reliability of LAN and further experimentation and ideas might be required to see if its quality can be improved.

We hope that this new direction would motivate further research in using attention-based mechanisms to effectively defend against adversarial examples. One possible idea is to improve the mask generator. Another is to use a different attention method: we build our work on Latent Attention Network but there might be other interpretability mechanisms that are better for adversary detection. We look forward to seeing more robust DNNs with the benefits of interpretability.

## 5 Appendix

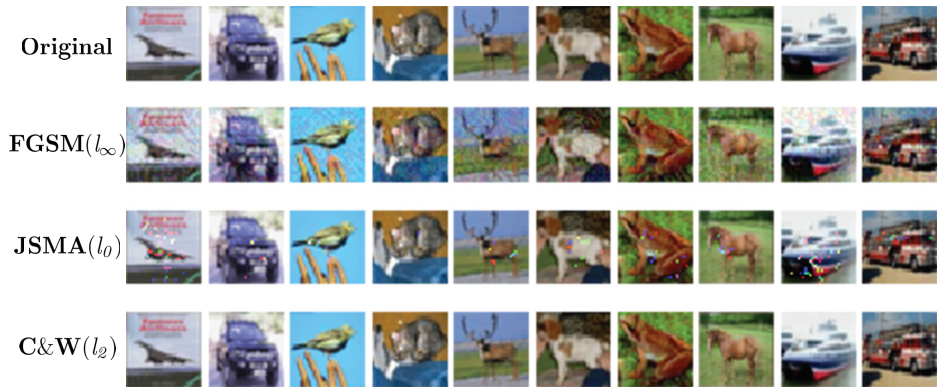

Figure 5: Original images and their adversarial counterparts generated by three attack models.

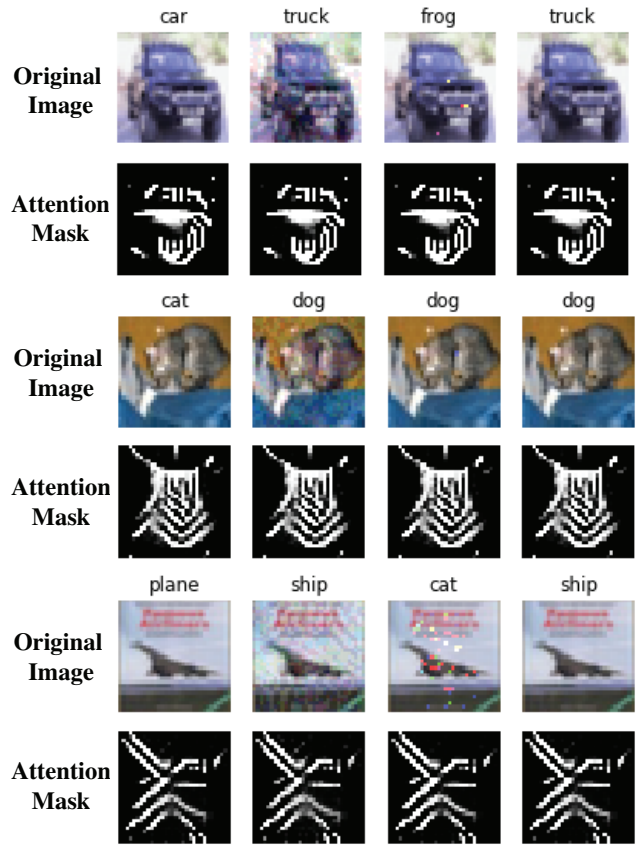

Figure 6: First row shows one benign image and its three adversarial examples by FGSM, JSMA, and C&W together with their classification results. The second row presents their corresponding attention masks. Visually, the masks look really similar to one another despite adversarial pertubations.

6